\documentclass{article} 
\usepackage[numbers,sort&compress]{natbib}
\usepackage[preprint]{nips_2021}
\usepackage[colorlinks=true,linkcolor=red,anchorcolor=red,citecolor=green,urlcolor=black]{hyperref}
\usepackage{url}
\usepackage{graphicx}
 \usepackage{subfigure} 
\usepackage{amsthm,amsmath,amssymb}
\usepackage{booktabs}
\usepackage{mathrsfs}
\usepackage{bm}
\usepackage{multirow}
\usepackage{cleveref}
\usepackage{color}
\usepackage{url}
\usepackage{wrapfig}
\usepackage{xcolor}

\title{MlTr: Multi-label Classification with Transformer}

\usepackage{array}
\makeatletter
\newcommand{\thickhline}{%
    \noalign {\ifnum 0=`}\fi \hrule height 1pt
    \futurelet \reserved@a \@xhline
}

\author{
  Xing Cheng\thanks{Interns at MMU, KuaiShou Inc.} \quad Hezheng Lin\footnotemark[1] \quad Xiangyu Wu\thanks{Corresponding author.} \quad Fan Yang
  \\\textbf{ Dong Shen\quad Zhongyuan Wang \quad Nian Shi\footnotemark[1] \quad Honglin Liu}  \\
  MMU \quad KuaiShou Inc.\\
  \texttt{\{chengxing03, linhezheng, wuxiangyu, yangfan\}@kuaishou.com} \\
  \texttt{\{shendong, wangzhongyuan, shinian, liuhonglin\}@kuaishou.com} \\
  
}

\begin{document}
\maketitle
\begin{abstract}
The task of multi-label image classification is to recognize all the object labels presented in an image. Though advancing for years, small objects, similar objects and objects with high conditional probability are still the main bottlenecks of previous convolutional neural network(CNN) based models, limited by convolutional kernels' representational capacity. Recent vision transformer networks utilize the self-attention mechanism to extract the feature of pixel granularity, which expresses richer local semantic information, while is insufficient for mining global spatial dependence. In this paper, we point out the three crucial problems that CNN-based methods encounter and explore the possibility of conducting specific transformer modules to settle them. We put forward a Multi-label Transformer architecture(MlTr) constructed with windows partitioning, in-window pixel attention, cross-window attention, particularly improving the performance of multi-label image classification tasks. The proposed MlTr shows state-of-the-art results on various prevalent multi-label datasets such as MS-COCO, Pascal-VOC, NUS-WIDE with \textbf{88.5\%}, \textbf{95.8\%}, \textbf{65.5\%} respectively. 
The code will be available soon at \url{https://github.com/starmemda/MlTr/}
\end{abstract}

\section{Introduction}
\label{intro}
\begin{figure}[h]
\begin{center}
\subfigure[Challenge of small objects]{\includegraphics[scale=0.4]{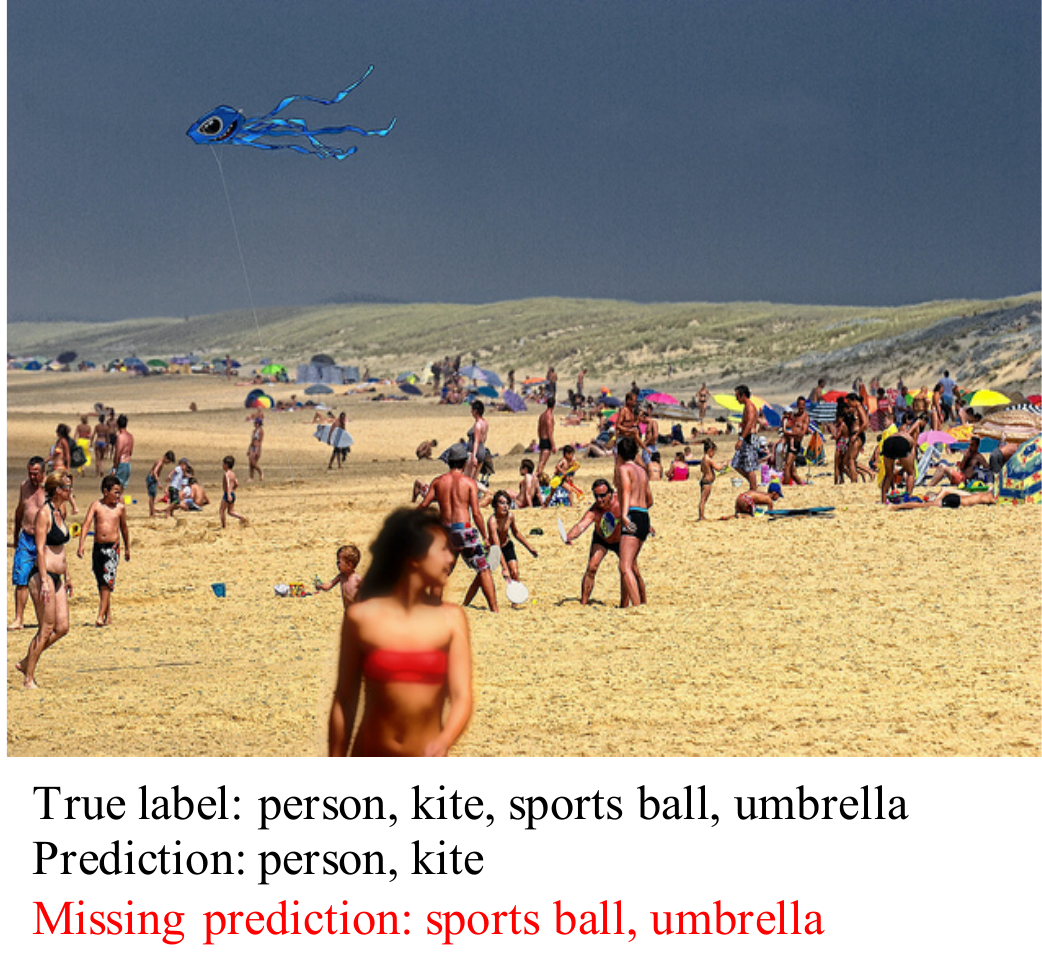}}
\subfigure[Challenge of similar objects]{\includegraphics[scale=0.4]{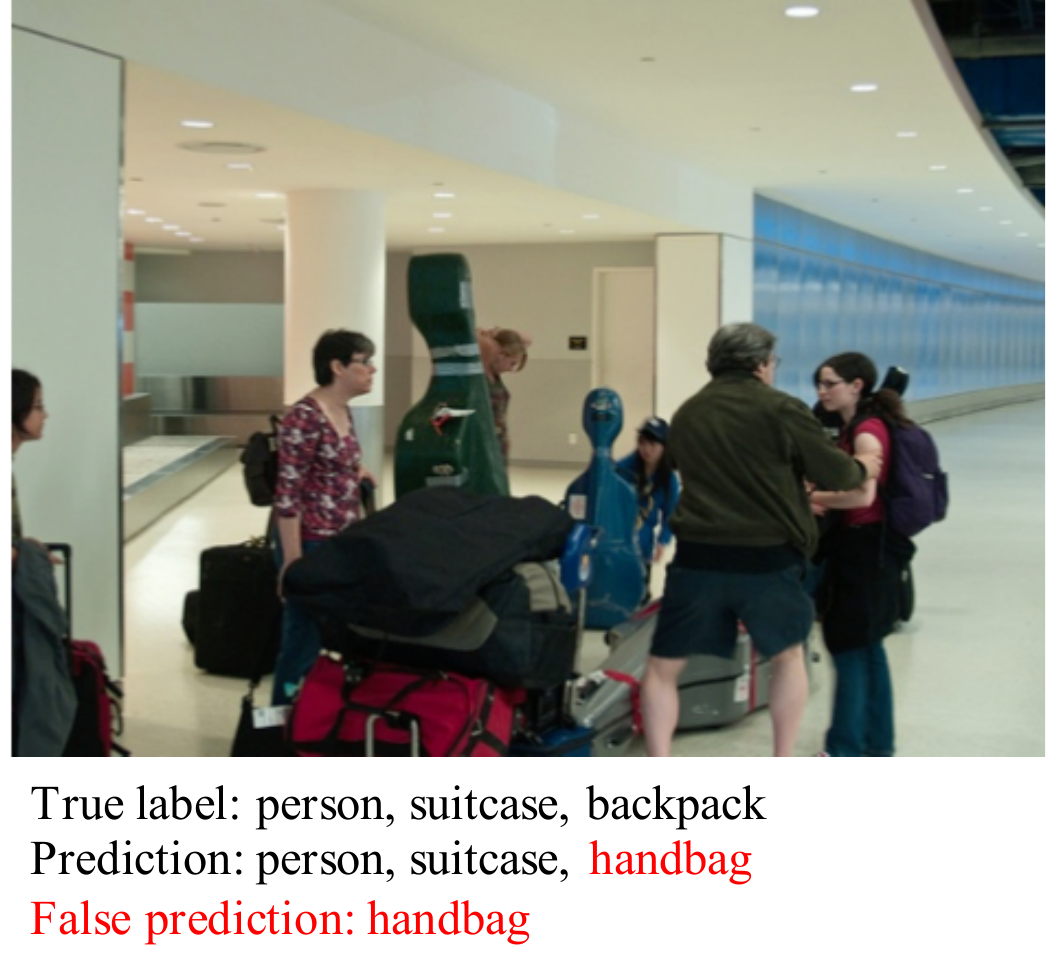} \label{similar}}
\subfigure[Confusion caused by objects occurring together frequently]{\includegraphics[scale=0.4]{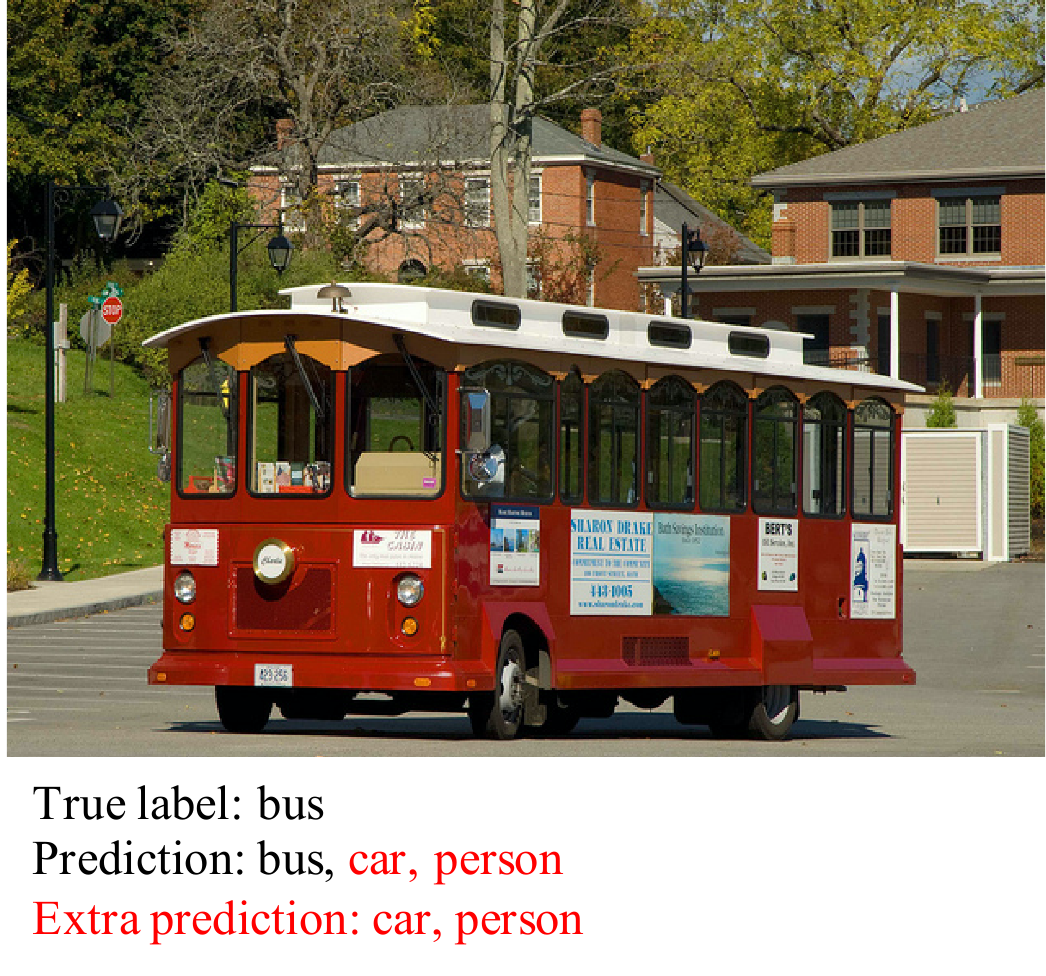}}
\end{center}
\caption{Three bad cases token from the ResNet101 baseline results}
\label{figure1}
\end{figure}

In general, multi-label image classification requires us to identify all the entities in one image and then label them accordingly. Compared to single-label image classification, such tasks have a wider range of applications in autonomous driving \cite{levinson2011towards}, medical diagnosis recognition \cite{ge2018chest}, and industrial-level image content understanding.\\
At present, mainstream multi-label classification schemes still take convolutional neural networks(CNN) as their primary frameworks. We analyzed several samples on MS-COCO \cite{lin2014microsoft} where the resnet101 \cite{he2016deep} model performed poorly and grouped them into three:

\begin{itemize}
\item Small objects. There are fewer pixels representing small objects which will disappear in the process of downsampling the feature map.
\item Different objects with high similarity. Similar categories have similar characteristics and are easily confused by model, e.g. backpack and handbag, as shown in Fig.\ref{similar}
\item Different objects that co-occur frequently. For instance, the condition that buses and cars often appear simultaneously, leading to inaccurate inner connections between their labels and features. It may misguide the model to distinguish a single bus as a car and a bus.
\end{itemize}

Since convolution is only computed in one local area at a time with the property of inductive bias, the local information is easily ignored after adequate times of convolution, resulting in CNN-based models' dissatisfying performances in multi-label classification. 

Transformer \cite{vaswani2017attention,devlin2018bert}, first created to excavate long-range dependencies among word embeddings in the field of Natural Language Processing(NLP), has also recently been found to be capable in various computer vision tasks. ViT \cite{dosovitskiy2020image} makes it possible for the transformer to replace convolutional neural networks' long-term domination in the field of computer vision. Unlike convolution's inherent inductive bias \cite{battaglia2018relational}, the transformer requires great amounts of data to learn long-range dependencies. Deit \cite{touvron2020training}, swin transformer \cite{liu2021swin} and Pit \cite{heo2021rethinking} introduce knowledge distillation, windows partitioning, and pooling operations to alleviate massive data requirement 
effectively. However, they share the basic pattern to partition the image into separate patches, project the patches into pixels and then conduct self-attention among pixels. This pattern limits the transformer's further application in multi-label image classification, where inferring small parts from the perspective of global spatial location relationships is of great importance. \\
In our work, we combine the pixel attention and the attention among windows to better excavate the transformer's activity in multi-label image classification. This new pattern of attention is what we call cross-window attention. In detail, We first project the input image into pixels within the feature map, and divide the pixels into window partitions, then alternate between pixel attention and 
cross window attention. With such a method, small objects can be detected by pixel attention and cross-window attention is capable of gathering overall information to make local conjecture more convincing. The object-aware token is added to our model's last layer to determine the number of labels, corresponding to the third problem. The proposed Multi-label Transformer architecture is named MlTr in short, and it achieves state-of-the-art performance on MS-COCO, Pascal-VOC, and NUS-WIDE with 88.5\%, 95.8\%, and 65.5\% respectively. Moreover, we show the visualization of the heatmaps in Fig.\ref{all_heatmap}, to elaborate that the proposed MlTr can indeed help each feature to capture implicit but crucial information globally. In summary, the contributions in this paper can be concluded as:
\begin{itemize}
\item Clearly pointing out the three crucial problems existing in the present multi-label image classification task.
\item Firstly proposing the transformer based network designed for multi-label classification, which surpasses previous SOTA methods on general benchmarks.
\item Offering a novel attention pattern of the transformer through putting forward cross-window attention.
\end{itemize}

\section{Related Work}
\label{rela}
\subsection{CNN Based Networks}
In the past many years, CNN models have brought the field of computer vision enormous progress. As general improvements of CNN were introduced to multi-label image classification tasks, there are some particular schemes being proposed according to their characteristics, mainly including Graph Convolution Networks(GCN) and weak supervision methods.\\
\textbf{GCN.}
In multi-label classification tasks, labels may share certain relationships as they normally co-occur in an image. A graph based on the conditional probability relation can be constructed as a prior to model the label dependencies. The papers based on GCN \cite{chen2019multi,you2020cross,chen2019learning,zhang2018multilabel,chen2018order,ye2020attention} utilize feature descriptions and correlation matrix with a graph convolutional neural network, and train a classifier to help the model infer. Although these series of methods can indeed bring some improvements, they are of little practical use. Mainly because they are not end-to-end, the graph relation among labels has to be extracted over training data, and the semantic priors have to be calculated by another NLP network in advance. Besides, the parameters of graph relation are the square of the number of categories, which is unaffordable when the number of categories is tremendous. But our proposed MlTr is end-to-end and can be extended for a large number of categories. \\
\textbf{Weak supervision methods.}
In the case of a large number of categories, even for a professional data labeling team, it is inevitable to miss some labels for few samples. Weak supervision \cite{lanchantin2020general,durand2019learning} mainly solves the problem of incomplete labeling that may exist in every sample, and it can be applied to multi-label classification. Their  principal means are to predict partial labels or calculate partial labels' loss during forward or backward propagation. 

\subsection{Vision Transformer Networks}
Since the paper \cite{dosovitskiy2020image}, the transformer has been utilized in many tasks of computer vision. Recent TNT, PiT and CvT \cite{han2021transformer,wu2021cvt,you2020cross,yuan2021tokens,heo2021rethinking,touvron2020training} have reached good performance in image classification. Swin Transformer \cite{liu2021swin,zhu2020deformable,carion2020end} even reached state-of-the-art in both object detection and image segmentation. As for video understanding, the transformer's self-attention mechanism is fully utilized to achieve unprecedented performance \cite{bertasius2021space,neimark2021video}. But primarily all transformer based networks notice only pixels' attention. For example, ViT directly conducts the transformer among global pixels, while TNT introduces local attention by alternating patch-level attention and pixel-level attention. Though swin transformer elaborated on increasing pixels' receptive field from local to global, they couldn't cast off the pixel based limitation. How to mine spatial dependence relations in a feature map directly and combine the pixel information, it's what the proposed MlTr in this work does. And we find it particularly suited to solving the three issues raised in Section \ref{intro}.

\subsection{Customized Loss Function}
In multi-label classification, the commonly used loss functions are bce loss and multi-label soft margin loss \cite{patterson2016coco} in the early stage, considering the problem of positive-negative imbalance \cite{lin2017focal}, asymmetric loss \cite{ben2020asymmetric} was proposed recently. It
combines the best of label smoothing \cite{muller2019does} and focal loss \cite{lin2017focal} that prevent overfitting and positive-negative imbalance by exploring an appropriate method to add margin in focal loss. It's so far the best loss function in multi-label classification, thus the loss optimization in this paper is based on asymmetric loss.

\section{Method: ML-Tr}
\label{method}
In this section, we illustrate the architecture of the proposed transformer based multi-label classifier and its composition of each submodule.

\subsection{MlTr}
\subsubsection{Overall Architecture}
\begin{figure}[h]
\begin{center}
\includegraphics[scale=0.30]{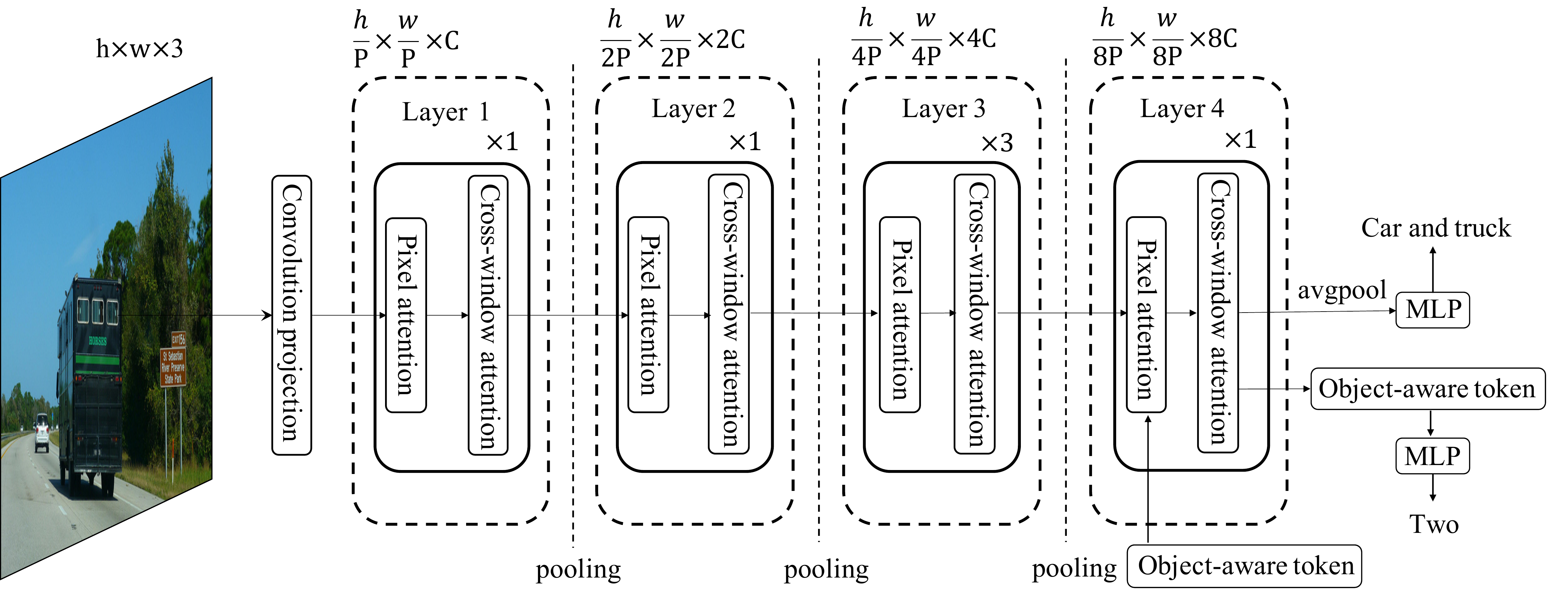}
\caption{Overall architecture of Multilabel Transformer(MlTr-s)}
\label{figure2}
\end{center}
\end{figure}
We design our model on scales: 224, 384, and 384, correspondingly named MlTr-s, MlTr-m, and MlTr-l with customized parameters respectively. Fig.\ref{figure2} shows our overall architecture. Absorbing the ideas of ResNet \cite{he2016deep} and Swin transformer \cite{liu2021swin}, we construct the network as a bottom-up model, which gradually increases the receptive field of each pixel from local to global. \\
Firstly, the input 2D image matrix is expanded into C feature maps of size $\frac{H}{P}\times\frac{W}{P}$ by convolution projection, where P denotes the patch size as paper \cite{vaswani2017attention}. Then, each pixel of the feature map is regarded as a token with dimension C and is input into multiple layers comprised of different blocks successively. Four layers of different sizes are designed, and between every two layers, space to depth module \cite{heo2021rethinking} is applied to reduce the feature size to $\frac{1}{4}$ and double the number of channels. In this way, we can gradually obtain more comprehensive information of the whole image through stacking the transformer module as typical convolution based networks do. Significantly, 
the object-aware token is stacked to the last layer to mine the number of distinct classes ever presented in one image. Eventually, the pixel tokens after average pooling \cite{boureau2010theoretical} are followed by multi-layer perceptron(MLP) to calculate optimized asymmetric loss \cite{ben2020asymmetric}, object-aware token is followed by MLP directly to calculate cross entropy loss \cite{jang2016categorical,de2005tutorial}.

\subsubsection{Attention Block}
\begin{figure*}[h]

\begin{center}
\subfigure[Illustration of pixel attention]{\includegraphics[scale=0.22]{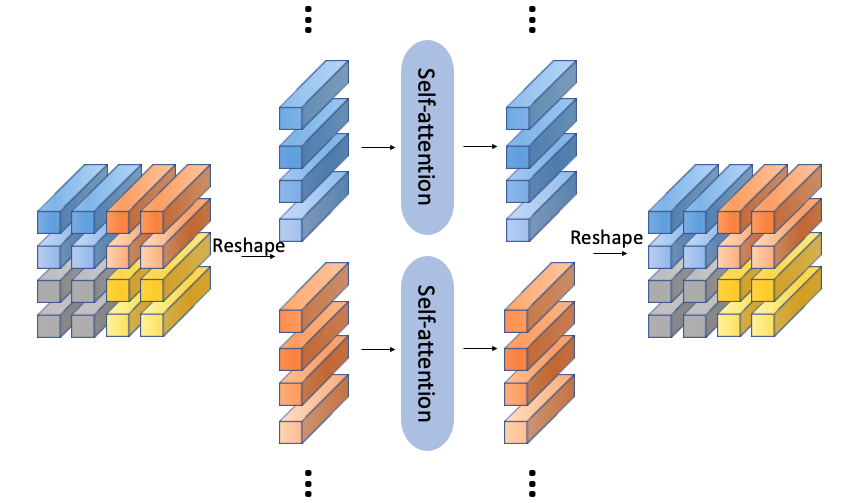}\label{figure3}}
\subfigure[Illustration of cross-window attention]{\includegraphics[scale=0.22]{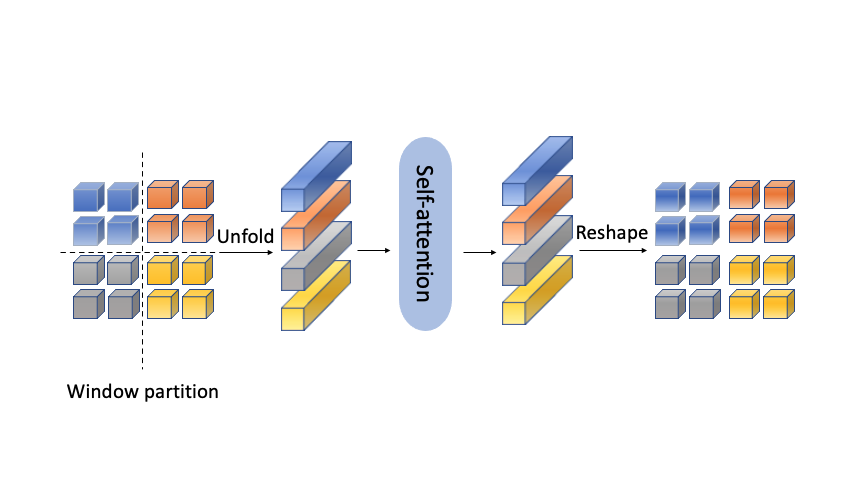}\label{figure4}}
\end{center}
\caption{A diagram of the MlTr block}
\end{figure*}
\begin{figure}[h]
\begin{center}
\includegraphics[scale=0.4]{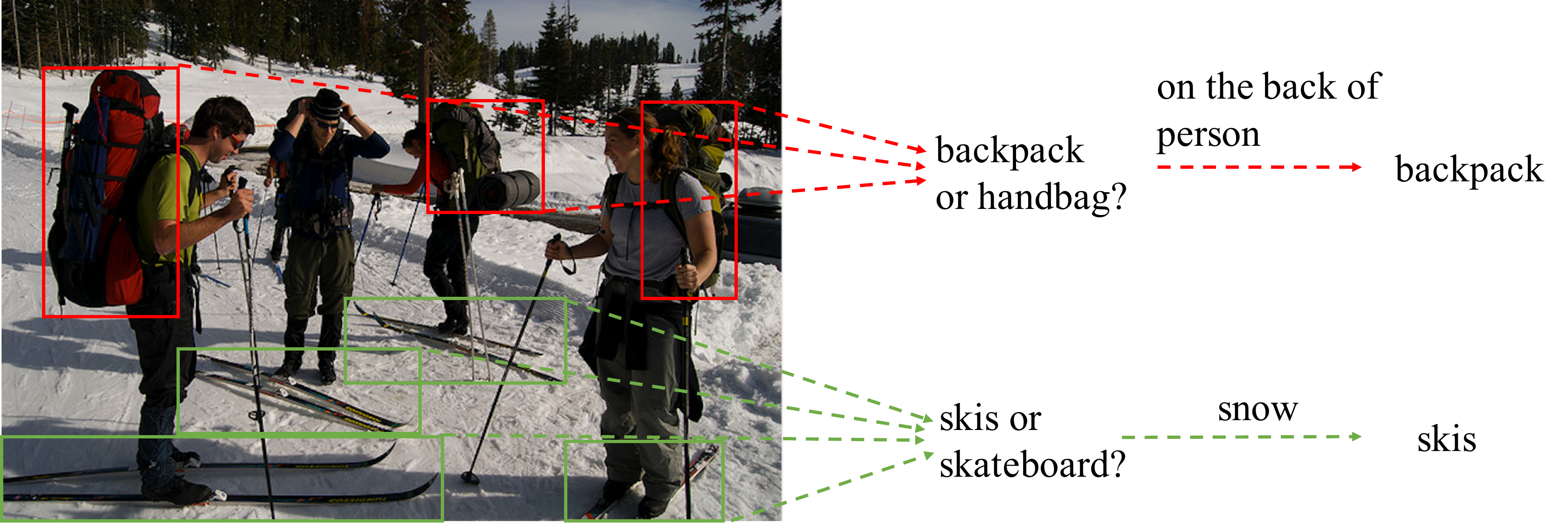}
\caption{The illustration of spatial dependence focused by cross-window attention}
\label{cross_illus}
\end{center}
\end{figure}

In this section, we illustrate the pixel attention and the cross-window attention that constitute the MlTr block together.\\\\
\textbf{Pixel Attention.} As Fig.\textcolor{red}{\ref{figure3}} shows, all pixels from the same position in the feature maps form a token and we set a certain window size to divide the feature map into different windows. Then self-attention \cite{vaswani2017attention} operation only works within windows. After utilizing the residual connection and multilayer perceptron, the calculation process can be expressed as:
\begin{align}
& \displaystyle x_{temp} = PA(LN(x^{l-1})) + x^{l-1}, \\
& \displaystyle x^l = MLP(LN(x_{temp})) + x_{temp} 
\end{align}
Where $x^{l-1}$ and $x^l$ denote the input and output of $l$-th attention module.$LN, PA$, and $MLP$ represent the layer normalization, pixel attention, and multilayer perceptron. Shift window operations are also necessary to expand pixels' receptive field \cite{liu2021swin}.\\\\
\textbf{Why Pixel Attention?} By partitioning windows, an artificial inductive bias was appended to the transformer. Based on the assumption that the most critical information for each object is primarily presented in and around the object, such inductive bias enables the network to converge efficiently on small datasets. And according to Section \ref{compute_cost}, the computational cost of pixel attention is far less than that of global attention adopted by ViT, TNT, and so on, which allows us to perform patch projecting in smaller patch size. Smaller patch size 
means denser image partitioning, retaining more local information, and achieving higher accuracy.\\\\
\textbf{Cross-window Attention.} Referred to Fig.\textcolor{red}{\ref{figure4}}. It turns different windows on one channel unfold into one-dimensional tokens whose length is $w_s^2$ where $w_s$ denotes the window size. Then, the subsequent self-attention operation was only conducted among the tokens derived from one feature map. Likewise:
\begin{align}
& x_{temp} = CWA(LN(x^{l-1})) + x^{l-1},\\
& x^l = MLP(LN(x_{temp})) + x_{temp}
\end{align}
Where $CWA$ means cross-windows attention.\\\\
\textbf{Why Cross-Window Attention?} 
In Fig.\ref{cross_illus}, with local pixel attention, backpack and skis are easily located. But backpack and handbag, skis and skateboard are so similar that the prediction score for them will aggregate around the threshold. Cross-window attention can exploit spatial dependence which may be location relations, scenes, and styles. Under the circumstances in Fig.\ref{cross_illus},  the backpacks are on the back of persons and the overall ground environment is snow, which will assist backpack and skis be inferred precisely. MlTr first makes a preliminary local judgment and then coordinates the global information, so as to obtain the prediction results with high confidence.
\subsubsection{Computational Cost Analysis}
\label{compute_cost}
Supposing the feature map of $C\times h\times w$ was partitioned by a window size of $w_s$. If we take the global attention strategy \cite{vaswani2017attention}, the computational cost is \cite{liu2021swin, han2021transformer}:
\begin{eqnarray}
 &FLOP_{GA} = 4hwC^2 + 2(hw)^2C \\
 &Param_{GA} \propto C^2     
\end{eqnarray}
Where GA denotes global attention. As for pixel attention and cross-window attention:
\begin{eqnarray}
 &FLOP_{PA} = 4hwC^2 + 2w_{s}^2hwC\\
 &Param_{PA} \propto C^2 \\
 &FLOP_{CWA} = 4hww_s^2C+2(\frac{hw}{w_s})^2C \\
 &Param_{CWA} \propto w_s^4 \ \
\end{eqnarray}

For computational complexity, the first two are both quadratic to channel number $C$, while the proposed spatial attention is linear. That's why pure depth attention networks can't afford deeper channels that may represent more features. But with cross-window attention, we are capable of alleviating this problem to some extent.\\
Otherwise, the parameter amount of pixel attention is irrelevant with channels $C$. Considering $C>>w_s^2$, adopting cross-window attention implies a significant reduction in the number of parameters.

\subsubsection{Architecture Variants}

We set up three MlTr models with different parameter amounts, computational complexity, and input resolutions to meet the needs of different application scenarios. Relevant variants are presented in Table \textcolor{red}{\ref{table5}}. $P$, $C$, and $w_s$ denote patch size, channels, and window size as Fig.\textcolor{red}{\ref{figure2}} shows. 

\begin{table*}[h]
\centering
\caption{Architecture Variants of MlTr.}
\label{table5}
\begin{tabular}{|c|c|c|c|c|c|}
\hline
\  & input resolution & $P$ & $C$ & $w_s$ &block numbers in four layer \\
\thickhline
MlTr-s &224 &4 &96 &7 &\{1,1,3,1\}  \\
\hline
MlTr-m &384 &4 &96  &12  &\{1,1,9,1\}  \\
\hline
MlTr-l &384 &4 &128  &12  &\{1,1,9,1\}  \\
\hline
\end{tabular}
\end{table*}

\subsection{Loss Optimizations}
\label{loss}
In this section, two optimizations towards loss are illustrated. 
\subsubsection{Arc Sigmoid}
The most widely used multi-label classification loss function, bce loss, is presented as follows:
\begin{eqnarray}
 &p_{ij} = \frac{1}{1+e^{x_{i}W_j^T+b_j}}\\
 &BCE_i = -\sum_{j}^{n}{y_{ij}}\log{p_{ij}}-\sum_{j}^{n}{(1-y_{ij})}\log(1-p_{ij})
\end{eqnarray}
where $x \in \mathbb{R}^{N\times{d}}$ denotes the features extracted by the model in $N$ samples and $d$ means the feature dimension which is often set to 2048 in the paper following \cite{he2016deep}, n is the total number of distinct categories. $i$ means $i$-th sample and $j$ means $j$-th class. So $W_j \in \mathbb{R}^{d}$ and $b_j \in \mathbb{R}^{d}$ denotes the $j$-th column of the weight of $W \in \mathbb{R}^{d\times{n}}$
and bias item in the final fully connected layer. True labels we wish the model to learn is y, $p_{ij}$ is the confidence calculated by the sigmoid function that the $i$-th sample belongs to $j$-th category.\\
We can infer that $W_j$ is the most typical feature of the $j$-th category, which is represented in the form of a vector in the d-dimensional hyperplane space. When the product of feature $x_i$ and $W_j$ learned by the model from an image reaches the maximum, we label the $i$-th sample as the $j$-th class. Referring to the paper \cite{deng2019arcface,liu2017sphereface,liu2016large}, it is more reasonable to describe the similarity among vectors by the angle on the hyperplane, so we put forward the arc sigmoid function which expresses as below:
\begin{eqnarray}
p_{ij} = \frac{1}{1+e^{s\cos\theta_{ij}}}
\label{arc_sig}
\end{eqnarray}
where $\theta_{ij} = \arccos\frac{x_{i}W_j^T}{||x_{i}||||W_j^T||}$ exactly means the angle between vector $x_{i}$ and $W_i^T$, s is a hyperparameter used to keep the predicted value within $[-s, s]$.
To alleviate the impact caused by samples with missing labels and positive-negative imbalance, we adopt asymmetric loss \cite{ben2020asymmetric} to replace bce loss:
\begin{eqnarray}
ASL_{i} = -\sum_{j}^{n}y_{ij}L_+-\sum_{j}^{n}(1-y_{ij})L_-
\end{eqnarray}
$L_+$ and $L_-$ are the positive and negative loss parts that can be calculated as:
\begin{eqnarray}
 &L_+ = (1-p_{ij})^{\gamma_{+}}\log(p_{ij}) \\
 &L_- = (p_{ij}-m)^{\gamma_-}\log(1-p_{ij}+m)
\end{eqnarray}
where $\gamma_{+}$, $\gamma_{-}$ and $m$ are the hyperparameters explored in \cite{ben2020asymmetric}.

\subsubsection{Object-aware Token Loss (OTL)}
Different from the labels' semantic priors extracted from the word2vec network in many papers \cite{chen2019multi,you2020cross,chen2019learning}, reinforced supervision can be summarized from the given labels, that is, the number of objects' distinct classes in an image. There are two ways to understand why such supervision works better, which requires no additional knowledge. On the one hand, it forces the model to extract the common features of the same class targets and enhances the generalization performance; on the other hand, it can make the model build a more robust corresponding projection between features and labels, especially for objects with high conditional probability. With the flexibility of transformer tokens, an object-aware token is stacked to MlTr's last layer to calculate the loss produced by the model's category numbers prediction. It can be derived from:
\begin{eqnarray}
OTL_{i} = -\log s_z
\end{eqnarray}
where $z=\sum_{j=0}^{n}y_{ij}$ represents the category numbers, and $s$ is the softmax output of the fully connected layer following object-aware token. In this way, the problem of quantity prediction is transformed into a classification task, $s_z$ means the predicted confidence that there are $z$ kinds of the object present in the image. \\
Eventually, the loss of $i$-th sample consists of two parts:
\begin{eqnarray}
L_{i} = ASL_i+OTL_i
\end{eqnarray}

\section{Experiments}
\label{result}
In this section, we employ the proposed MlTr method to enforce plentiful experiments on several datasets universally token for multi-label classification. Then compare the results with the previous architectures and prove the effectiveness.
\subsection{Datasets, Metrics, and Implementation Details}
\textbf{Datasets.} We pretrain MlTr on ImageNet-22K \cite{deng2009imagenet} which contains 14.2 million images from 22K classes and then fine-tune on multi-label classification benchmarks such as MS-COCO \cite{lin2014microsoft}, Pascal-VOC \cite{everingham2010pascal}, and NUS-WIDE \cite{chua2009nus}. They consist of 122218 images, 9963 images, and 220000 images from 80, 20, and 81 classes respectively. \\\\
\textbf{Metrics.} Under the premise that label the sample with the class if its predicted score in one class is greater than a threshold  (e.g. 0.5). The metric we're most interested in is mean average precision (mAP) \cite{chen2019multi}, some additional indicator reported on previous work such as average per-class precision (CP), recall (CR), F1 (CF1), and average overall precision (OP), recall (OR), F1(OF1) are requisite, too.\\\\
\textbf{Implementation Details.} With mainstream fine-tune strategies like Adam \cite{loshchilov2017decoupled}, asymmetric loss \cite{ben2020asymmetric}, one-circle \cite{smith2018disciplined} learning rate scheduler as tresnet \cite{ridnik2021tresnet}, we carry out our experiment. $s, m, \gamma_+$, and $\gamma_-$ in Section \ref{loss} is set to be 8, 0.05, 0, and 4. Every image is transformed by cutout \cite{devries2017improved}, randaugment when training, and the initial learning rate is 1e-4. The result will be shown after 200 epochs. More detailed configurations can be seen in our open source code.

\subsection{Results on Benchmark and Comparisons with State-of-the-Arts}
\textbf{MS-COCO.} 
The results on the MS-COCO dataset are shown in Table \textcolor{red}{\ref{table1}}. As we can see, our smallest design is 3.8\% higher than resnet101 \cite{he2016deep} when they are both pretrained on imagenet1k with 224 resolution input. The medium-computational MlTr-m version with 384 resolution can achieve the same performance as the previous best method TresNet-l, which even needs a more intricate scale with 448 input size. When it comes to more complicated architecture like MlTr-l, the mAP can reach 88.5, it is 1.9\% higher than the previous top one. Overall, the transformer based network we design can deliver more accurate results at a lower resolution input, and the large one MlTr-l demonstrates the ability to challenge higher precision.

\begin{table*}
\caption{Comparison of various metrics with previous state-of-the-art on MS-COCO. The default unit is \%.The proposed MlTr-s, MlTr-m and MlTr-l use resolutions of 224,384,384 respectively. 1k and 22k denote the model is pretrained on imagenet1k and imagenet22k.}
\label{table1}
\centering
\begin{tabular}{|c||c|c|c|c|c|c|c|c|c|}
\hline
\  & Resolution &mAP & CP & CR & CF1 & OP & OR & OF1 &Params(M) \\
\hline
\hline
ResNet-101 \cite{he2016deep} &$224\times224$ &78.3 &80.2 &66.7 &72.8  &83.9 &70.8 &76.8 &45\\
ML-GCN \cite{chen2019multi} &$448\times448$ &83.0 &85.1 &72.0 &78.0 &85.8 &75.4 &80.3   &46\\
SSGRL \cite{chen2019learning}  &$448\times448$ &83.8 &89.9 &68.5 &76.8 &91.3 &70.8 &79.9   &-\\
KGGR \cite{chen2020knowledge}   &$448\times448$ &84.3 &85.6 &72.7 &78.6 &87.1 &75.6 &80.9   &-\\
C-Tran \cite{lanchantin2020general} &$448\times448$ &85.1 &86.3 &74.3 &79.9 &87.7 &76.5 &81.7   &-\\
Tresnet-l \cite{ben2020asymmetric} &$448\times448$ &86.6 &87.2 &76.4 &81.4 &88.2 &79.2 &81.8  &55\\
\hline
\hline
MlTr-s (1k) &$224\times224$ &81.9 &80.7 &71.5 &75.2 &81.4 &76.3 &78.1   &33\\
MlTr-s (22k) &$224\times224$ &83.9 &82.8 &75.5 &77.3 &83.0 &78.5 &79.9 &33\\
MlTr-m (22k)&$384\times384$ &86.8 &84.0 &80.1 &81.7 &84.6 &82.5 &83.5   &62\\
MlTr-l (22k)&$384\times384$ &\textbf{88.5} &\textbf{86.0} &\textbf{81.4} &\textbf{83.3} &\textbf{86.5} &\textbf{83.4} &\textbf{84.9}   &108\\
\hline
\end{tabular}
\end{table*}

\textbf{Pascal-VOC.}
To be fair, we only take the models pretrained on ImageNet for comparison. As Table \textcolor{red}{\ref{table2}} shows, MlTr-l reaches a new state-of-the-art, it demonstrates that although the Pascal-VOC dataset is relatively small, it doesn't prevent MlTr from performing better. This also validates MLTr's strong capacity for transfer learning. 

\begin{minipage}{\textwidth}
\begin{minipage}[t]{0.48\textwidth}
\makeatletter\def\@captype{table}
\caption{Comparison of mAP with previous excellent methods on Pascal-VOC dataset.}
\label{table2}
\centering
\begin{tabular}{c|c}
\hline
\  & mAP \\
\hline
Atten-Reinforce \cite{chen2018recurrent} &92.0 \\
SSGRL \cite{chen2019learning} &93.4 \\
ML-GCN \cite{chen2019multi} &94.0\\
Tresnet-l \cite{ben2020asymmetric} &94.6\\
\hline
MlTr-l &\textbf{95.8}\\
\hline
\end{tabular}
\end{minipage}
\begin{minipage}[t]{0.48\textwidth}
\makeatletter\def\@captype{table}
\caption{Comparison of mAP, CF1, and OF1 with previous method on NUS-WIDE dataset.}
\label{table3}
\centering
\begin{tabular}{c|c|c|c}
\hline
\  & mAP & CF1 & OF1\\
\hline
MS-CMA \cite{you2020cross} &61.4 &60.5 &73.8 \\
SRN \cite{zhu2017learning} &62.0 &58.5 &73.4 \\
ICME \cite{chen2019multi} &62.8 &60.7 &74.1 \\
Tresnet-l \cite{ben2020asymmetric} &65.2 &63.6 &75.0 \\
\hline
MlTr-l &\textbf{66.3} &\textbf{65.0} &\textbf{75.8}   \\
\hline
\end{tabular}
\end{minipage}
\end{minipage}

\textbf{NUS-WIDE.}
With the proposed MlTr-l, mAP reaches 66.3\% on NUS-WIDE as Table \textcolor{red}{\ref{table3}} shows, CF1 and OF1 have also been promoted,  all of these indicate MlTr's better generalization performance.

\subsection{Ablation Studies}
\begin{wraptable}{r}{5cm}
\centering
\caption{Ablation study of MlTr-s on the benchmark MS-COCO.}
\label{Ablation1}
\begin{tabular}{c|c}
\hline
\  & mAP  \\
\thickhline
fixed window &79.2 \\
\hline
+position embedding \cite{liu2021swin} &79.6(+0.4) \\
+shift window \cite{liu2021swin} &80.5(+0.9) \\
+cross-window attention &81.5(+1.0)\\
+arc sigmoid &81.7(+0.2)\\
+object-aware token &81.9(+0.2) \\
\hline
\end{tabular}
\end{wraptable}
\textbf{Main methods.} To prove the efficiency of the refinements put forward in our work, we remove them in order and test the mAP on MS-COCO's validation set. The relevant result is shown in Table \textcolor{red}{\ref{Ablation1}}, position embedding, shift window, cross-window attention, arc sigmoid, and object-aware token produce an increase of 0.4, 0.9, 1.0, 0.2, and 0.2 separately. As expected, cross-window attention enables the model exploit the spatial dependence and significantly improve the performance.

\textbf{Arc sigmoid.} Referred to Eq.(\ref{arc_sig}), the proposed arc sigmoid connects the predicted score and the similarity described by cosine value between embeddings. Usually $cosine\theta \in [-1,1]$ \cite{deng2019arcface}, which indicates the result concentrates on 0.5 if feeding the $cosine\theta$ into sigmoid function directly. In Table \ref{s}, we assign scaling factors $s$ as 4, 8, 16, 32, and it turns out that it performs best when $s$ is equal to 8.

\begin{table*}[h]
\centering
\caption{Ablation tests with different s values in Eq.(\ref{arc_sig}). All the results are obtained by MlTr-s on MS-COCO.}
\label{s}
\begin{tabular}{|c|c|c|c|c|}
\hline
s  & 4 & 8 & 16 & 32 \\
\hline
mAP &81.7 &81.9 &81.5 &81.2 \\
\hline
\end{tabular}
\end{table*}

\textbf{Small Objects.}
According to the bounding box provided by MS-COCO and its criterion for small, medium, and large objects, we divide all labels of images in the validation dataset into these three and calculate their mAP as $AP_s$, $AP_m$, and $AP_l$. In the case that two entities of different size is the same category, we boil the label down to the larger one. The results of ResNet101 and MlTr-s are compared in Table \ref{small}. As it shows, the average precision for small objects($AP_s$) increases by nearly 10 points, which is more obvious than $AP_m$ and $AP_l$.

\begin{table*}[h]
\centering
\caption{The comparisons in three different size objects.}
\label{small}
\begin{tabular}{|c|c|c|c|c|}
\hline
\  & $AP_s$ & $AP_m$ & $AP_l$ & $mAP$ \\
\thickhline
ResNet101 \cite{he2016deep} &68.8 &83.8 &92.6 &82.2 \\
\hline
MlTr-m &77.6 &90.2 &95.9  &88.5\\
\hline
\end{tabular}
\end{table*}

\begin{figure}[h]
\begin{center}
\includegraphics[scale=0.27]{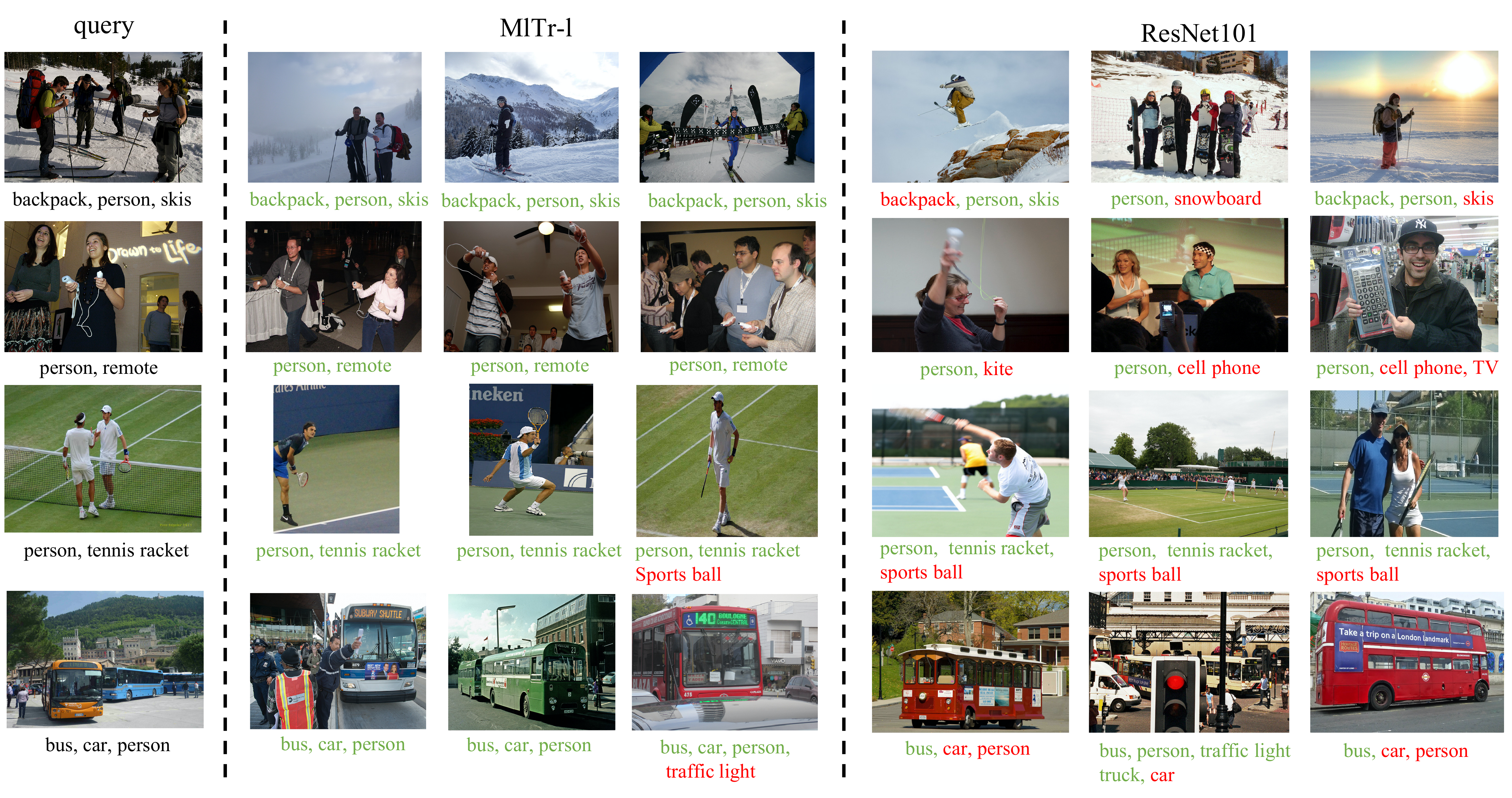}
\caption{Three picked out images from Top-5 returned ones with the query images. The leftmost column is the query images, our proposed MlTr-l returns with the left three, while the right three are based on ResNet101 baseline.}
\label{bijiao}
\end{center}
\end{figure}

\textbf{Similar Objects And Objects with High Conditional Probability.}
The effectiveness of our approach towards similar objects and objects with high probability is visualized in this section. In Fig.\ref{bijiao}, MlTr-l and ResNet101 with the accuracy described in the last part are compared by t-SNE \cite{van2008visualizing}.\\
In the first two examples, MlTr performs significantly better in distinguishing similar couples such as skis and snowboard, backpack and clothing, remote and cell phone. For the last two rows display, due to the high probability of tennis rackets and sports balls, cars, and buses appearing in the same image, ResNet will identify any suspicious dot on the tennis court as a sports ball, and label the image 
of a bus with the car label. This is because the local feature extracted by ResNet isn't convincing enough, it learns an implicit graph relation to converge, which is the local optimum. With the object-aware token, the proposed MlTr is forced to wipe off the graph relation and maintain a more meaningful semantic topology.

\section{Conclusion}
In this paper, we firstly analyse the three bottlenecks in multi-label image classification and then propose a novel transformer based architecture majoring in them. By effectively combining the pixel attention and cross-window attention, it makes a preliminary judgment of the target locally and then matches the relevant features globally. Besides, two loss optimizations have also been implemented to correct the feature distribution. The proposed MlTr is the first transformer framework to solve the problem of multi-label classification and reaches the state-of-the-art in various multi-label benchmarks, it further suggests the broad prospects of transformer thriving in the field of computer vision.

\bibliographystyle{unsrt}
\bibliography{myref}

\appendix

\section{The Visualization of Heatmaps}
\begin{figure}[h]
\begin{center}
\subfigure[The visualization of heatmap. The first column: original image, the second to the fourth ones: heatmap reflecting the attention to different objects.\label{heatmap} ]{\includegraphics[scale=0.22]{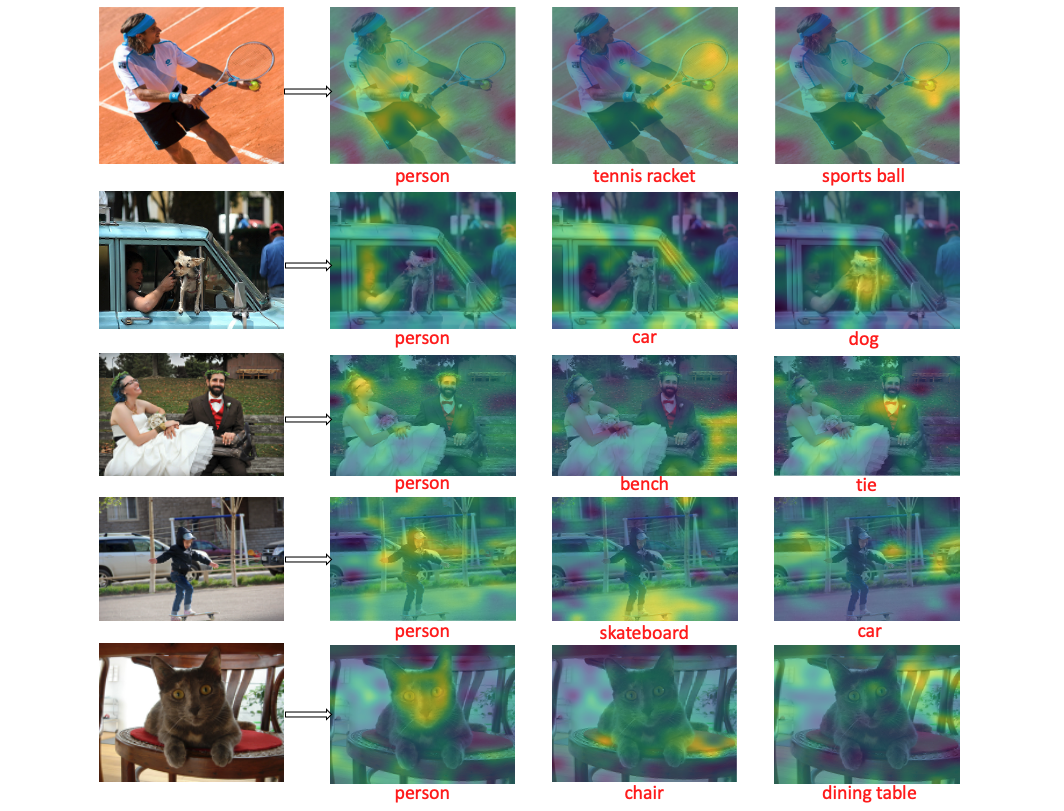}}
\subfigure[Two examples of spatial attention when recognizing the tennis rocket or sports ball. The first column: original image, the second column: heatmaps calculated from a same channel output, the third one: heatmaps taking the second column pictures' greatest concern regions as highlight.\label{spatial_man_women} ]{\includegraphics[scale=0.15]{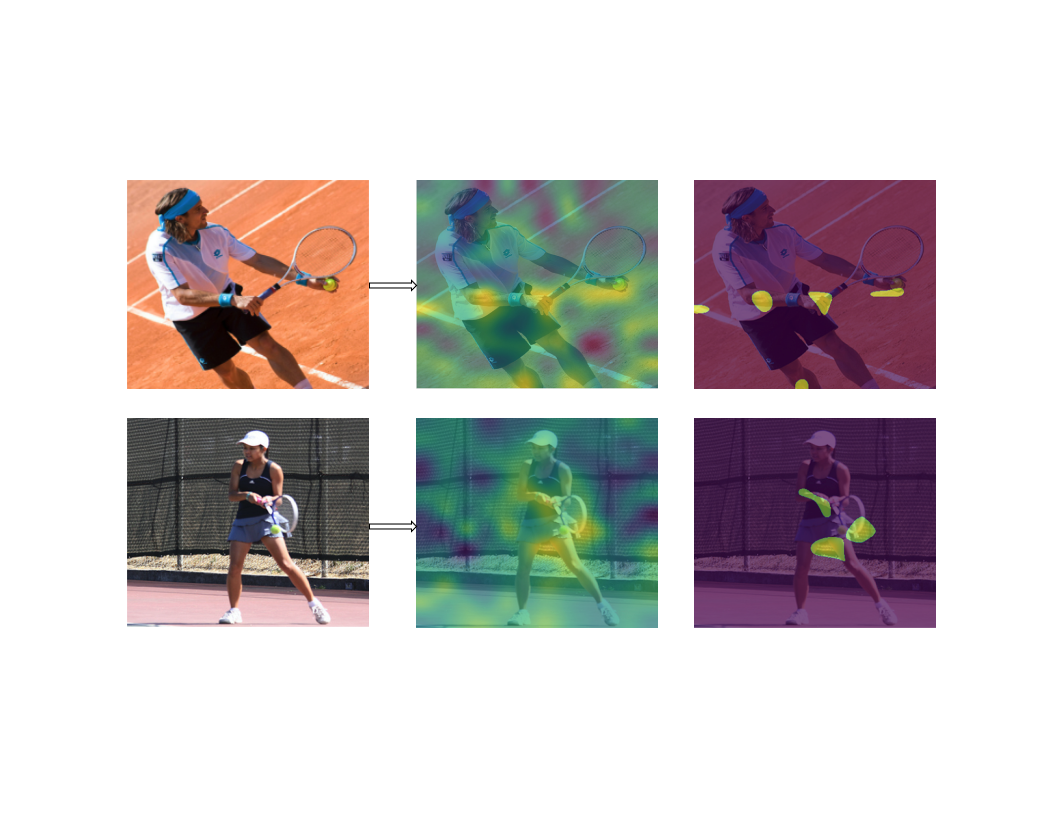}}
\caption{The comparison of heatmaps.}
\label{all_heatmap}
\end{center}
\end{figure}

In this section, we visualize the heatmap \cite{woo2018cbam} reflecting the focus of each channel derived from the last layer's feature map. The specific approach is to upsample(bilinear) each feature map to the original input size. After normalization, draw the heatmap and the initial image in a same picture. For the tennis playing picture as Fig.\textcolor{red}{\ref{heatmap}} shows, feature maps from different channels recognize clues of different objects.\\
Interestingly, we found that some feature maps employ rich spatial information when identifying tennis balls. Take Fig.\textcolor{red}{\ref{spatial_man_women}} as an example, with our cross-window attention, when discerning the sports ball, our model will simultaneously focus on the person's arm, hand and leg, who is preparing to kick the ball. To prove such ingenious spatial attention is no coincidence, we select another tennis playing image and feed it into the network with the same parameters, and draw the feature map from the same channel output. It turns out that the heatmap also 
focuses on the right hand, the thigh and the tennis racket. \\
The reason for such a phenomenon may be that, when the model detection preliminarily determines that the small target may be a tennis ball, the entity, scene or action related to tennis ball will be searched by cross-window attention in the global scope. If the relevant pixels are found, it can be more certain that the small target is a tennis ball, which is similar to the reasoning of human sometimes.

\end{document}